\documentclass[preprint,12pt, a4paper]{elsarticle}

\usepackage{amssymb}

\usepackage{lineno}

\usepackage{float}
\restylefloat{table}

\usepackage{booktabs}
\usepackage{subcaption}
\usepackage{algorithm}
\usepackage[noend]{algpseudocode}
\usepackage{float}
\usepackage{xcolor}
\usepackage{listings}
\usepackage{wrapfig}
\usepackage{courier}
\usepackage{enumitem}
\definecolor{codegreen}{rgb}{0,0.6,0}
\definecolor{codegray}{rgb}{0.5,0.5,0.5}
\definecolor{codepurple}{rgb}{0.58,0,0.82}
\definecolor{backcolour}{rgb}{0.95,0.95,0.92}
\PassOptionsToPackage{hyphens}{url}\usepackage{hyperref}
\renewcommand{\texttt}[1]{%
  \begingroup
  \ttfamily
  \begingroup\lccode`~=`/\lowercase{\endgroup\def~}{/\discretionary{}{}{}}%
  \begingroup\lccode`~=`[\lowercase{\endgroup\def~}{[\discretionary{}{}{}}%
  \begingroup\lccode`~=`,\lowercase{\endgroup\def~}{,\discretionary{}{}{}}%
  \begingroup\lccode`~=`.\lowercase{\endgroup\def~}{.\discretionary{}{}{}}%
  \catcode`/=\active\catcode`[=\active\catcode`.=\active\catcode`,=\active
  \scantokens{#1\noexpand}%
  \endgroup
}
\usepackage{adjustbox}
\usepackage{array}
\newcolumntype{R}[2]{%
    >{\adjustbox{angle=#1,lap=\width-(#2)}\bgroup}%
    l%
    <{\egroup}%
}
\usepackage{makecell}

\lstdefinestyle{mystyle}{
    backgroundcolor=\color{backcolour},   
    commentstyle=\color{codegreen},
    keywordstyle=\color{magenta},
    numberstyle=\tiny\color{codegray},
    stringstyle=\color{codepurple},
    basicstyle=\footnotesize\ttfamily,
    breakatwhitespace=false,         
    breaklines=true,                 
    captionpos=b,                    
    keepspaces=true,                 
    numbers=none,                    
    numbersep=5pt,                  
    showspaces=false,                
    showstringspaces=false,
    showtabs=false,                  
    tabsize=2
}

\journal{SoftwareX}

\begin{document}

\begin{frontmatter}



\title{Sherpa: Robust Hyperparameter Optimization for Machine Learning}


\author[1]{Lars Hertel\corref{cor1}}
\ead{lhertel@uci.edu}
\author[2]{Julian Collado}
\ead{colladou@uci.edu}
\author[3]{Peter Sadowski}
\ead{peter.sadowski@hawaii.edu}
\author[2]{Jordan Ott}
\ead{jott1@uci.edu}
\author[2]{Pierre Baldi}
\ead{pfbaldi@ics.uci.edu}

\cortext[cor1]{Corresponding author}
\address[1]{Department of Statistics
Donald Bren School of Information and Computer Sciences
University of California, Irvine
Bren Hall 2019
Irvine, CA 92697-1250, USA}
\address[2]{Department of Computer Science
Donald Bren School of Information and Computer Sciences
University of California, Irvine
3019 Donald Bren Hall
Irvine, CA 92697-3435, USA}
\address[3]{Information and Computer Science
University of Hawai'i at M\~anoa, 1680 East-West Rd, Honolulu, HI 96822, USA}

\begin{abstract}
Sherpa is a hyperparameter optimization library for machine learning models. It is specifically designed for problems with computationally expensive, iterative function evaluations, such as the hyperparameter tuning of deep neural networks. With Sherpa, scientists can quickly optimize hyperparameters using a variety of powerful and interchangeable algorithms.
Sherpa can be run on either a single machine or in parallel on a cluster. Finally, an interactive dashboard enables users to view the progress of models as they are trained, cancel trials, and explore which hyperparameter combinations are working best.
Sherpa empowers machine learning practitioners by automating the more tedious aspects of model tuning. Its source code and documentation are available at \url{https://github.com/sherpa-ai/sherpa}.
\end{abstract}

\begin{keyword}
Hyperparameter Optimization \sep Machine Learning \sep Deep Neural Networks



\end{keyword}

\end{frontmatter}



\section{Motivation and significance}
\label{sec:motivation-and-significance}

Hyperparameters are tuning parameters of machine learning models. Hyperparameter optimization refers to the process of choosing optimal hyperparameters for a machine learning model. This optimization is crucial to obtain optimal performance from the machine learning model. Since hyperparameters cannot be directly learned from the training data, their optimization is often a process of trial and error conducted manually by the researcher. There are two problems with the trial and error approach. Firstly, it is time consuming and can take days or even weeks of the researcher's attention. Secondly, it is dependent on the researcher's ability to interpret results and choose good hyperparameter settings. These limitations lead to a large need to automate this process. Sherpa is a software that addresses this need. 


Existing hyperparameter optimization software can be divided into bayesian optimization software, bandit and evolutionary algorithm software, framework specific software, and all-round software. Software that implements bayesian optimization started with \href{https://github.com/automl/SMAC3}{SMAC} \citep{hutter2011sequential}, \href{https://github.com/HIPS/Spearmint}{Spearmint} \citep{snoek2012practical}, and \href{https://github.com/hyperopt/hyperopt}{HyperOpt} \citep{bergstra2013hyperopt}. More recent software in this regime has been  \href{https://github.com/SheffieldML/GPyOpt}{GPyOpt} \citep{gpyopt2016}, \href{https://github.com/automl/RoBO}{RoBo} \citep{klein-bayesopt17},  \href{https://github.com/dragonfly/dragonfly}{DragonFly} \citep{kandasamy2019tuning}, \href{https://github.com/wujian16/Cornell-MOE}{Cornell-MOE} \citep{wu2016parallel, wu2017bayesian}, and \href{https://github.com/mlr-org/mlrMBO}{mlrMBO} \citep{mlrMBO}. These software packages have high quality, stand-alone bayesian optimization implementations, often with unique twists. However, most of these do not provide infrastructure for parallel training.


As an alternative to bayesian optimization, multi-armed bandits and evolutionary algorithms have recently become popular. \href{https://github.com/automl/HpBandSter}{HpBandSter} implements Hyperband \citep{li2017hyperband} and BOHB \citep{falkner2018bohb}, \href{https://github.com/MattKleinsmith/pbt}{Pbt} implements Population Based Training \citep{jaderberg2017population}, \href{https://github.com/CMA-ES/pycma}{PyCMA} implements CMA-ES \citep{igel2006computational}, and \href{https://github.com/EpistasisLab/tpot}{TPot} \citep{Olson2016EvoBio, OlsonGECCO2016} provides hyperparameter search via genetic programming.


A number of framework specific libraries have also been proposed. \href{https://github.com/automl/autoweka}{Auto-Weka} \citep{kotthoff2017auto} and \href{https://github.com/automl/auto-sklearn}{Auto-Sklearn} \citep{feurer2015efficient} focus on WEKA \citep{holmes1994weka} and Scikit-learn \citep{pedregosa2011scikit}, respectively. Furthermore, a number of packages have been proposed for the machine learning framework Keras \citep{chollet2015keras}. \href{https://github.com/maxpumperla/hyperas}{Hyperas}, \href{
https://github.com/keras-team/autokeras}{Auto-Keras} \citep{jin2019auto}, \href{https://github.com/autonomio/talos}{Talos}, \href{https://github.com/Avsecz/kopt}{Kopt}, and \href{https://github.com/joeddav/devol}{HORD} each provide hyperparameter optimization specifically for Keras. These libraries make it easy to get started due to their tight integration with the machine learning framework. However, researchers will inevitably run into limitations when a different machine learning framework is needed.


Lastly, a number of implementations aim at being framework agnostic and also support multiple optimization algorithms. Table~\ref{tab:comparison-table} shows a detailed comparison of these "all-round" packages to Sherpa. Note that we excluded Google Vizier \citep{golovin2017google} and similar frameworks from other cloud computing providers since these are not free to use. 


\newcommand{\No}{{\color{red}No}} 
\newcommand{\Yes}{{\color{blue}Yes}} 
\newcommand*\rot{\multicolumn{1}{R{45}{1em}}}

\begin{table}[]
\centering
\footnotesize
\begin{tabular}{lcccccc}
\thead{Software}  & \thead{Distributed} & \thead{Visualizations} & \thead{Bayesian-\\Optimization} & \thead{Evolutionary} & \thead{Bandit/\\Early-stopping} \\
Sherpa 
&\Yes         &\Yes       &\Yes      &\Yes          &\Yes\\
\href{https://github.com/tobegit3hub/advisor}{Advisor}
&\Yes         &\No       &\Yes      &\Yes          &\Yes\\
\href{https://github.com/AIworx-Labs/chocolate}{Chocolate}
&\Yes         &\No       &\Yes      &\Yes          &\No\\
\href{https://github.com/williamFalcon/test-tube}{Test-Tube}\citep{Falcon2017}
&\Yes         &\No       &\No      &\No          &\No\\
\href{https://github.com/ray-project/ray/tree/master/python/ray/tune}{Ray-Tune}\citep{liaw2018tune}
&\Yes         &\No       &\No      &\Yes          &\Yes\\
\href{https://github.com/pfnet/optuna}{Optuna}\citep{akiba2019optuna}
&\Yes         &\Yes       &\Yes      &\No          &\Yes\\
\href{https://github.com/HDI-Project/BTB}{BTB} \citep{Laura:2018}
&\No         &\No       &\Yes      &\No          &\Yes\\
\end{tabular}
\caption{Feature comparison of hyperparameter optimization frameworks. \textit{Bayesian optimization}, \textit{evolutionary}, and \textit{bandit/early-stopping} refer to the support of hyperparameter optimization algorithms based on these methods.\label{tab:comparison-table}}
\end{table}

Sherpa is already being used in a wide variety of applications such as machine learning methods \cite{sadowski2018neural}, solid state physics \cite{cao2019convolutional}, particle physics \cite{baldi2019improved}, medical image analysis \cite{ritter2019hyperparameter}, and cyber security\cite{langford2019robust}. Due to the fact that the number of machine learning applications is growing rapidly we can expect there to be a growing need for hyperparameter optimization software such as Sherpa.

\section{Software Description}
\label{sec:software-description}

\subsection{Hyperparameter Optimization}
\label{sec:hyperparameter-optimization}
We begin by laying out the components of a hyperparameter optimization. Consider the training of a machine learning model. A user has a \textit{model} that is being trained with \textit{data}. Before training there are hyperparameters that need to be set. At the end of the training we obtain an \textit{objective} value.

This workflow can be illustrated via the training of a neural network. The \textit{model} is a neural network. The \textit{data} are images that the neural network is trained on. The \textit{hyperparameter setting} is the number of hidden layers of the neural network. The \textit{objective} is the prediction accuracy on a hold-out dataset obtained at the end of training.

For automated hyperparameter optimization we also need hyperparameter \textit{ranges}, a \textit{results} table,  and a hyperparameter optimization \textit{algorithm}. The hyperparameter ranges define what values each hyperparameter is allowed to take. The results store hyperparameter settings and their associated objective value. Finally, the algorithm takes results and ranges and produces a new suggestion for a hyperparameter setting. We refer to this suggestion as a \textit{trial}.

For the neural network example the hyperparameter range might be 1, 2, 3, or 4 hidden layers. We might have previous results that 1 corresponds to 80\% accuracy and 3 to 90\% accuracy. The algorithm might then produce a new trial with 4 hidden layers. After training the neural network with 4 hidden layers we find it achieves 88\% accuracy and add this to the results. Then the next trial is suggested.



\subsection{Components}
\begin{figure}[H]
    \centering
    \includegraphics[trim=11cm 14cm 5cm 4cm, clip, width=0.7\textwidth]{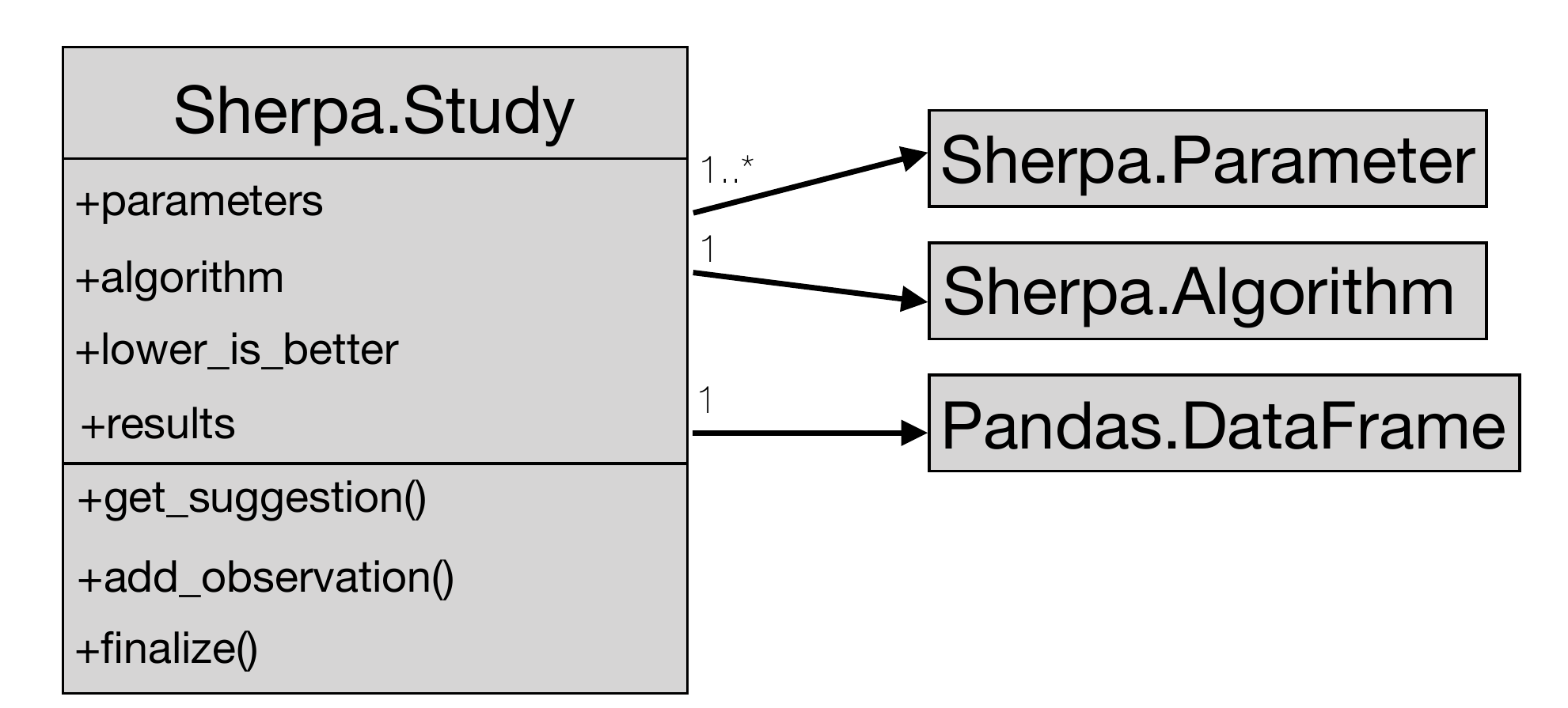}
    \caption{Diagram showing Sherpa's \texttt{Study} class.}
    \label{fig:sherpa-study}
\end{figure}

We now describe how Sherpa implements the components described in Section~\ref{sec:hyperparameter-optimization}. Sherpa implements hyperparameter ranges as \texttt{sherpa.Parameter} objects. The \textit{algorithm} is implemented as a \texttt{sherpa.algorithms.Algorithm} object. A list of hyperparameter ranges and an algorithm are combined to create a \texttt{sherpa.Study} (Figure~\ref{fig:sherpa-study}). The study stores the \textit{results}. Trials are implemented as \texttt{sherpa.Trial} objects.


Sherpa implements two user interfaces. We will refer to the two interfaces as \textit{API mode} and \textit{parallel mode}.

\subsection{API Mode}
In API mode the user interacts with the \texttt{Study} object. Given a study \textit{s}:
\begin{enumerate}
    \item A new trial of name \texttt{t} is obtained by calling \texttt{s.get\_suggestion()} or by iterating over the study (e.g. \texttt{for t in s}).
    \item First, \texttt{t.parameters} is used to initialize and train a machine learning model. Then \texttt{s.add\_observation(t, objective=o)} is called to add objective \texttt{o} for trial \texttt{t}. Invalid observations are automatically excluded from the results.
    \item Finally, \texttt{s.finalize(t)} informs Sherpa that the model training is finished.
\end{enumerate}
 Interacting with the Study class is easy. It also requires minimal setup. The limitation in API mode is that it cannot evaluate trials in parallel.




\subsection{Parallel Mode}
\begin{figure}[H]
    \centering
    \includegraphics[trim=0cm 8cm 1cm 2cm, clip, width=\textwidth]{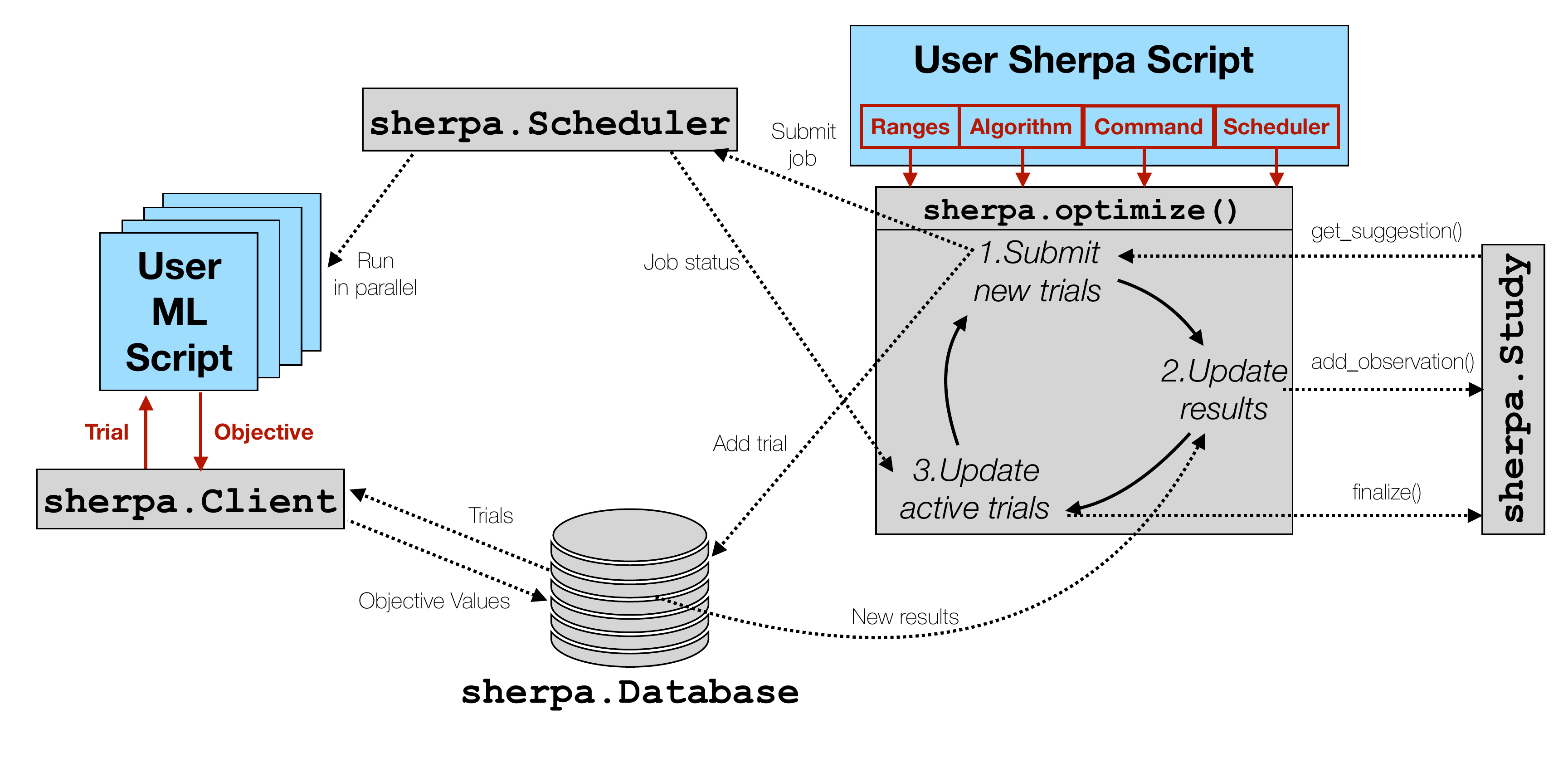}
    \caption{Architecture diagram for parallel hyperparameter optimization in Sherpa. The user only interacts with Sherpa via the solid red arrows, everything else happens internally.}
    \label{fig:sherpa-architecture}
\end{figure}

In \textit{parallel-mode} multiple trials can be evaluated in parallel. The user provides two scripts: a \textit{server script} and a \textit{machine learning (ML) script}.  The \textit{server script} defines the hyperparameter ranges, the algorithm, the job scheduler, and the command to execute the machine learning script. The optimization starts by calling \texttt{sherpa.optimize}.\\
In the \textit{machine learning script} the user trains the machine learning model given some hyperparameters and adds the resulting objective value to Sherpa. Using a \texttt{sherpa.Client} called \texttt{c} a trial \texttt{t} is obtained by calling \texttt{c.get\_trial()}. To add observations \texttt{c.send\_metrics(trial=t, objective=o)} is used.\\
Internally, \texttt{sherpa.optimize} runs a loop that uses the Study class. Figure~\ref{fig:sherpa-architecture} illustrates the \textit{parallel-mode} architecture.
\begin{enumerate}
    \item The loop submits new trials if resources are available by submitting a job to the scheduler. Furthermore, the new trials are added to a database. From there they can be retrieved by the client.
    \item The loop updates results by querying the database for new results.
    \item Finally, the loop checks whether jobs have finished. This means resources are free again. In addition, the corresponding trials can be finalized.
\end{enumerate}
 If the user's machine learning script does not submit an objective value such as when it crashed, Sherpa continues with the next trial.

\section{Software Functionalities}
\label{sec:software-functionalities}

\subsection{Available Hyperparameter Types}
Sherpa supports four hyperparameter types:
\begin{itemize}
    \item \texttt{sherpa.Continuous}
    \item \texttt{sherpa.Discrete}
    \item \texttt{sherpa.Choice}
    \item \texttt{sherpa.Ordinal}.
\end{itemize}
These correspond to a range of floats, a range of integers, an unordered categorical variable, and an ordered categorical variable, respectively. Each parameter has \texttt{name} and \texttt{range} arguments. The range expects a list defining lower and upper bound for continuous and discrete variables.
For choice and ordinal variables the range expects the categories.


\subsection{Diversity of Algorithms\label{sec:diversity-of-algorithms}}
Sherpa aims to help researchers at various stages in their model development. For this reason, it provides a choice of hyperparameter tuning algorithms. The following optimization algorithms are currently supported.
\begin{itemize}
    \item \texttt{sherpa.algorithms.RandomSearch}:\\Random Search \cite{bergstra2012random} samples hyperparameter settings uniformly from the specified ranges. It is a robust algorithm  because it explores the space uniformly. Furthermore, with the dashboard the user can make their own inference on the results.
    \item \texttt{sherpa.algorithms.GridSearch}:\\Grid Search follows a grid over the hyperparameter space and evaluates all combinations. It is useful to systematically explore one or two hyperparameters. It is not recommended for more than two hyperparameters.
    \item \texttt{sherpa.algorithms.bayesian\_optimization.GPyOpt}:\\Bayesian optimization is a model-based search. For each trial it picks the most promising hyperparameter setting based on prior results. Sherpa's implementation wraps the package GPyOpt \citep{gpyopt2016}.
    \item \texttt{sherpa.algorithms.successive\_halving.SuccessiveHalving}:\\Asynchronous Successive Halving (ASHA) \citep{li2018massively} is a hyperparameter optimization algorithm based on multi-armed bandits. It allows the efficient exploration of a large hyperparameter space. This is accomplished by the early stopping of unpromising trials.
    \item \texttt{sherpa.algorithms.PopulationBasedTraining}:\\Population-based Training (PBT) \citep{jaderberg2017population} is an evolutionary algorithm. The algorithm jointly optimizes a population of models and their hyperparameters. This is achieved by adjusting hyperparameters during training. It is particularly suited for neural network training hyperparameters such as learning rate, weight decay, or batch size.
    \item \texttt{sherpa.algorithms.LocalSearch}:\\Local Search is a heuristic algorithm. It starts with a seed hyperparameter setting. During optimization it randomly perturbs one hyperparameter at a time. If a setting improves on the seed then it becomes the new seed. This algorithm is particularly useful if the user already has a well performing hyperparameter setting.
\end{itemize}
All implemented algorithms allow parallel evaluation and can be used with all available parameter types. An empirical comparison of the algorithms can be found in the documentation\footnote{\url{https://parameter-sherpa.readthedocs.io/en/latest/algorithms/algorithms.html}}.

\subsection{Accounting for Random Variation}
Sherpa can account for variation via the \texttt{Repeat} algorithm. The objective value of a model may vary between training runs. Reasons for this can be random initialization or stochastic training. The \texttt{Repeat} algorithm runs each hyperparameter setting multiple times. Thus variation can be taken into account when analyzing results.

\subsection{Visualization Dashboard}
\begin{figure}[H]
  \centering
  \includegraphics[width=\linewidth]{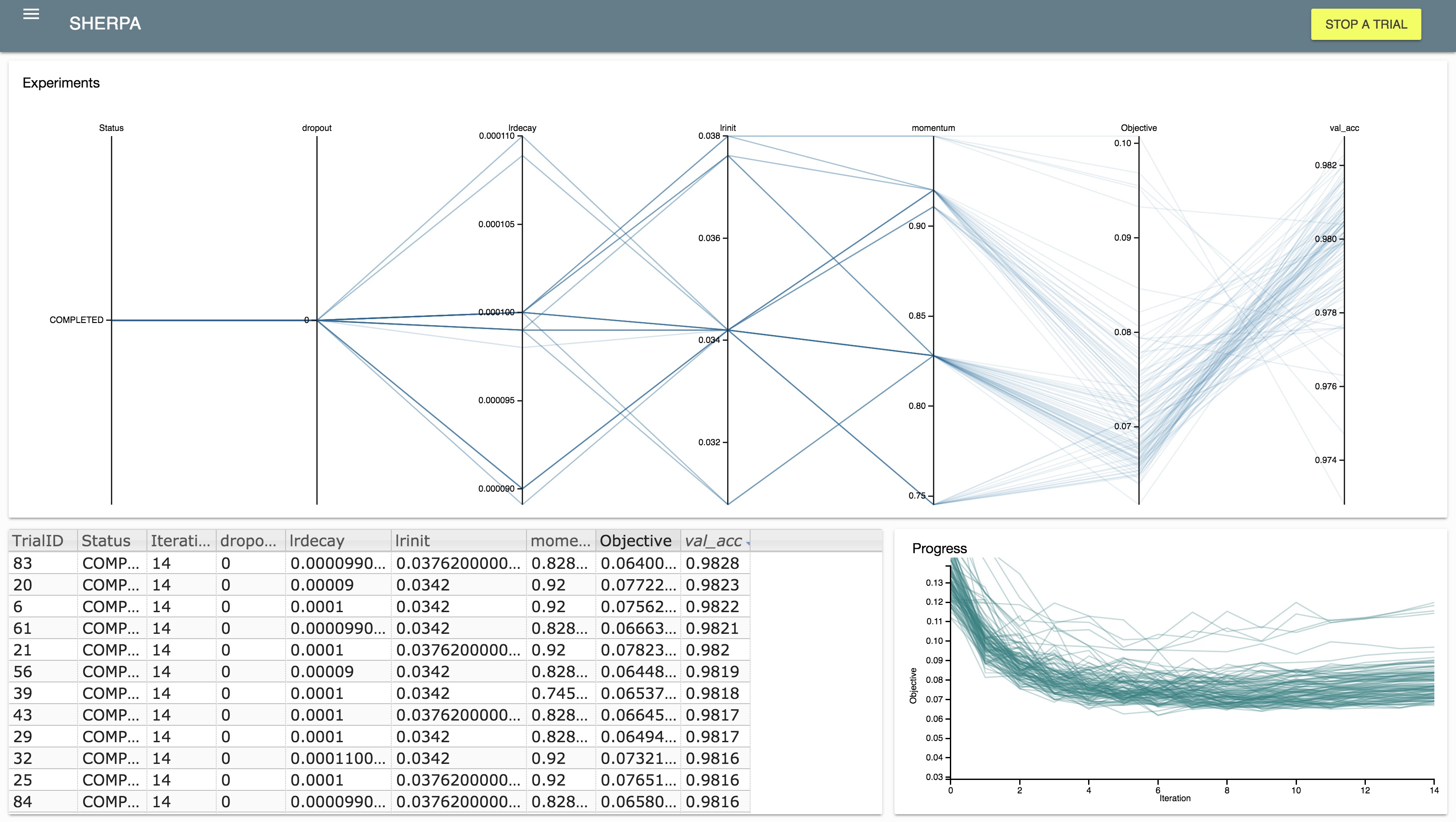}
  \caption{The dashboard provides a parallel coordinates plot (top) and a table of finished trials (bottom left). Trials in progress are shown via a progress line chart (bottom right). Figure recommended to be viewed as PDF and via zooming in.}
  \label{fig:dashboard}
\end{figure}

Sherpa provides an interactive web-based dashboard. It allows the user to monitor progress of the hyperparameter optimization in real time. Figure~\ref{fig:dashboard} shows a screenshot of the dashboard.

At the top of the dashboard is a parallel coordinates plot \citep{inselberg1987parallel,hauser2002angular}. It allows exploration of relationships between hyperparameter settings and objective values (Figure~\ref{fig:dashboard} top). Each vertical axis corresponds to a hyperparameter or the objective. The axes can be brushed over to select subsets of trials. The plot is implemented using the D3.js parallel-coordinates library by \citet{parallel-coordinates}.
At the bottom right is a line chart. It shows objective values against training iteration (Figure~\ref{fig:dashboard} bottom right). This chart allows to monitor training progress of each trial. It is also useful to analyze whether a trial's training converged.
At the bottom left is a table of all completed trials  (Figure~\ref{fig:dashboard} bottom left). Hovering over trials in the table highlights the corresponding lines in the plots.
Finally, the dashboard has a stopping button (Figure~\ref{fig:dashboard} top right corner). This allows the user to cancel the training for unpromising trials. 

The dashboard runs automatically during a hyperparameter optimization. It can be accessed in a web-browser via a link provided by Sherpa. The dashboard is useful to quickly evaluate questions such as: 
\begin{itemize}
    \item Are the selected hyperparameter ranges appropriate?
    \item Is training unstable for some hyperparameter settings?
    \item Does a particular hyperparameter have little impact on the performance of the machine learning algorithm?
    \item Are the best observed hyperparameter settings consistent?
\end{itemize}
Based on these observations the user can refine the hyperparameter ranges or choose a different algorithm, if appropriate.

\subsection{Scaling up with a Cluster}

In \textit{parallel mode} Sherpa can run parallel evaluations. A job scheduler is responsible for running the user's machine learning script. The following job schedulers are implemented.
\begin{itemize}
    \item The \texttt{LocalScheduler} evaluates parallel trials on the same computation node. This scheduler is useful for running on multiple local CPU cores or GPUs. It has a simple resource handler for GPU allocation (see Figure~\ref{fig:mnist-parallel-mode} for an example).
    \item The \texttt{SGEScheduler} uses Sun Grid Engine (SGE) \citep{gentzsch2001sun}. Submission arguments and an environment profile can be specified via arguments to the scheduler. 
    \item The \texttt{SLURMScheduler} is based on SLURM \citep{yoo2003slurm}. Its interface is similar to the \texttt{SGEScheduler}.
\end{itemize}
Concurrency between workers is handled via MongoDB, a NoSQL database program. Parallel mode expects that MongoDB is installed on the system.

\section{Illustrative Examples}
\label{sec:illustrative-examples}

\subsection{Handwritten Digits Classification with a Neural Network}
The following is an example of a Sherpa hyperparameter optimization. It uses the MNIST handwritten digits dataset \cite{deng2012mnist}. A Keras neural network is used to classify the digits. The neural network has one hidden layer and a softmax output. The hyperparameters are the learning rate of the Adam \cite{kingma2014adam} optimizer, the number of hidden units, and the hidden layer activation function. The search is first conducted using Sherpa's API mode. After that we show the same example using Sherpa's parallel mode.

\subsubsection{API Mode\label{sec:api-mode-example}}
Figure~\ref{fig:mnist-api-mode} shows the hyperparameter optimization in Sherpa's API mode. The script starts with imports and loading of the MNIST dataset. Next, the hyperparameters \textit{learning\_rate}, \textit{num\_units}, and \textit{activation} are defined. These refer to the Adam learning rate, number of hidden layer units, and hidden layer activation function, respectively. As optimization algorithm the \textit{GPyOpt} algorithm is chosen. Hyperparameter ranges and algorithm are combined via the \textit{Study}. The \texttt{lower\_is\_better} flag indicates that lower objective values are not better. This is because we will be maximizing the classification accuracy.
After that a for-loop iterates over the study. The for-loop yields a trial at each iteration. A Keras model is instantiated using the hyperparameter settings. The Keras model is iteratively trained and evaluated via an inner for-loop. We add an observation for each iteration and use \texttt{finalize} after the training is finished. Note that we pass the loss as context to \texttt{add\_observation}. The context accepts a dictionary with any additional metrics that the user wants to record.
Code to replicate this example is available as a Jupyter notebook\footnote{\url{https://github.com/sherpa-ai/sherpa/blob/master/examples/keras_mnist_mlp.ipynb}} and on Google Colab\footnote{\url{https://colab.research.google.com/drive/1I19R1GfKPjlgNdHlxJwNC4PitvySsdon}}. A video tutorial is also available on YouTube\footnote{\url{https://youtu.be/-exnF3uv0Ws}}. Tutorials using the Successive Halving and Population Based Training algorithms are also available\footnote{\url{https://github.com/sherpa-ai/sherpa/blob/master/examples/keras_mnist_mlp_successive_halving.ipynb}}\footnote{\url{https://github.com/sherpa-ai/sherpa/blob/master/examples/keras_mnist_mlp_population_based_training.ipynb}}.

\begin{figure}[H]
    \begin{lstlisting}[language=Python, style=mystyle]
import sherpa
import sherpa.algorithms.bayesian_optimization as bayesian_optimization
import keras
from keras.models import Sequential
from keras.layers import Dense, Flatten
from keras.datasets import mnist
from keras.optimizers import Adam
epochs = 15
(x_train, y_train), (x_test, y_test) = mnist.load_data()
x_train, x_test = x_train/255.0, x_test/255.0

# Sherpa setup
parameters = [sherpa.Continuous('learning_rate', [1e-4, 1e-2]),
              sherpa.Discrete('num_units', [32, 128]),
              sherpa.Choice('activation',
                            ['relu', 'tanh', 'sigmoid'])]
algorithm = bayesian_optimization.GPyOpt(max_num_trials=50)
study = sherpa.Study(parameters=parameters,
                     algorithm=algorithm,
                     lower_is_better=False)

for trial in study:
    lr = trial.parameters['learning_rate']
    num_units = trial.parameters['num_units']
    act = trial.parameters['activation']

    # Create model
    model = Sequential([Flatten(input_shape=(28, 28)),
                        Dense(num_units, activation=act),
                        Dense(10, activation='softmax')])
    optimizer = Adam(lr=lr)
    model.compile(loss='sparse_categorical_crossentropy',
                  optimizer=optimizer,
                  metrics=['accuracy'])

    # Train model
    for i in range(epochs):
        model.fit(x_train, y_train)
        loss, accuracy = model.evaluate(x_test, y_test)
        study.add_observation(trial=trial, iteration=i,
                              objective=accuracy,
                              context={'loss': loss})
    study.finalize(trial=trial)
\end{lstlisting}
  \caption{An example showing how to tune the hyperparameters of a neural network on the MNIST dataset using Sherpa in API mode.}
  \label{fig:mnist-api-mode}
\end{figure}

\subsubsection{Parallel Mode\label{sec:parallel-mode-example}}
We now show the same hyperparameter optimization using Sherpa's parallel mode. Figure~\ref{fig:mnist-parallel-mode} (top) shows the server script. First, the hyperparameters and search algorithm are defined. This time we also define a \texttt{LocalScheduler} instance. Hyperparameters, algorithm, and scheduler are passed to the \texttt{sherpa.optimize} function. We also pass a command "python trial.py". The command indicates how to execute the user's machine learning script. Furthermore, the argument \texttt{max\_concurrent=2} indicates that two evaluations will be running at a time.
Figure~\ref{fig:mnist-parallel-mode} (bottom) shows the machine learning script. First, we set environment variables for GPU configuration. Next we create a \textit{Client}. To obtain hyperparameters we call the client's \texttt{get\_trial} method. Furthermore, during training we call the client's \texttt{send\_metrics} method. This replaces \texttt{add\_observation} in parallel mode. Also, in parallel mode no \textit{finalize} call is needed.

\begin{figure}[H]
  \begin{center}
\begin{lstlisting}[language=Python, style=mystyle]
import sherpa
import sherpa.algorithms.bayesian_optimization as bayesian_optimization
from sherpa.schedulers import LocalScheduler
params = [sherpa.Continuous('learning_rate', [1e-4, 1e-2]),
              sherpa.Discrete('num_units', [32, 128]),
              sherpa.Choice('activation',
                            ['relu', 'tanh', 'sigmoid'])]
alg = bayesian_optimization.GPyOpt(max_num_trials=50)
sched = LocalScheduler(resources=[0,1])
sherpa.optimize(parameters=params, algorithm=alg,
                scheduler=sched, lower_is_better=False,
                command='python trial.py', max_concurrent=2)
\end{lstlisting}

\begin{lstlisting}[language=Python, style=mystyle]
import sherpa
import os
GPU_ID = os.environ['SHERPA_RESOURCE']
os.environ['CUDA_VISIBLE_DEVICES'] = GPU_ID
import keras
from keras.models import Sequential
from keras.layers import Dense, Flatten
from keras.datasets import mnist
from keras.optimizers import Adam
epochs = 15
(x_train, y_train), (x_test, y_test) = mnist.load_data()
x_train, x_test = x_train/255.0, x_test/255.0
# Sherpa client
client = sherpa.Client()
trial = client.get_trial()
lr = trial.parameters['learning_rate']
num_units = trial.parameters['num_units']
act = trial.parameters['activation']
# Create model
model = Sequential([Flatten(input_shape=(28, 28)),
                    Dense(num_units, activation=act),
                    Dense(10, activation='softmax')])
optimizer = Adam(lr=lr)
model.compile(loss='sparse_categorical_crossentropy',
              optimizer=optimizer,
              metrics=['accuracy'])
# Train model
for i in range(epochs):
    model.fit(x_train, y_train)
    loss, accuracy = model.evaluate(x_test, y_test)
    client.send_metrics(trial=trial, iteration=i,
                        objective=accuracy,
                        context={'loss': loss})
\end{lstlisting}
\end{center}
\caption{A code listing showing how to use Sherpa in parallel mode to tune the hyperparameters of a neural network trained on the handwritten digits dataset MNIST. The top code listing shows the server-script. The bottom listing shows the trial-script.}
\label{fig:mnist-parallel-mode}
\end{figure}



\subsection{Deep learning for Cloud Resolving Models\label{sec:case-study}}
\subsubsection{Introduction}
The following illustrates an example of a Sherpa hyperparameter optimization in the field of climate modeling, specifically cloud resolving models (CRM). We apply Sherpa to optimize the deep neural network (DNN) of \citet{rasp2018deep}.

The input to the model is a 94-dimensional vector. Features include temperature, humidity, meridional wind, surface pressure, incoming solar radiation, sensible heat flux, and latent heat flux. 
The output of the DNN is a 65-dimensional vector. It is composed of the sum of the CRM and radiative heating rates, the CRM moistening rate, the net radiative fluxes at the top of the atmosphere and surface of the earth, and the observed precipitation.

\subsubsection{General Hyperparameter Optimization\label{sec:general-search}}
Initially a random search was conducted on the following hyperparameters: batch normalization \citep{ioffe2015batch}, dropout \citep{srivastava2014dropout, baldi2013understanding}, Leaky ReLU coefficient \citep{agostinelli2014learning}, learning rate, nodes per hidden layer, number of hidden layers. The parameter ranges were chosen to encompass the parameters specified in \cite{rasp2018deep}. From the dashboard (Figure~\ref{fig:case_study_dashboard}) we identify that the best performing configurations have low dropout, leaky ReLU coefficients mostly around 0.3 or larger, and learning rates mostly near 0.002. The majority of good models have 8 layers and batch normalization. However, the number of units does not seem to have a large impact. The hyperparameter ranges and best configuration are provided in Tables~\ref{hyperparameter-table} and~\ref{tab:bestconfig} in the appendix.

\subsubsection{Optimization of the Learning Rate Schedule\label{sec:secondary-search}}
An additional search was conducted to fine-tune the DNN training hyperparameters. Specifically, the initial learning rate and the learning rate decay were optimized. The range of initial learning rate values was $\pm 10^{-4}$ of the best value from Section~\ref{sec:general-search}. The range of learning rate decay factors was $0.5$ to $1$. The learning rate gets multiplied by this factor after every epoch to produce a new learning rate. In comparison, the model in \citet{rasp2018deep} uses a decay factor of approximately $0.58$. The remaining hyperparameters were set to the best configuration from Section~\ref{sec:general-search}. A total of 50 trials were evaluated via random search. The best learning rate was found to be $0.001196$. The best decay value was found as $0.843784$. The overall optimal hyperparameter setting is shown in Table~\ref{tab:bestconfig} of the supplementary materials.

\subsubsection{Results}
We compare the model found by Sherpa to the model from \citet{rasp2018deep} via $R^2$ plots (Figure~\ref{fig:case_study_results}). The $R^2$ plots show the coefficient of determination at different pressures and latitudes. We find that the Sherpa model consistently outperforms the comparison model. In particular, it is able to perform for latitudes for which the prior model fails. Figure~\ref{fig:case_study_loss_plot} shows that the Sherpa model's loss reduces further after the \citet{rasp2018deep} model has converged. This is the result of the learning rate fine-tuning from Section~\ref{sec:secondary-search}.

\section{Impact}
\label{sec:impact}



Machine learning is used to ever larger extends in the scientific community. Nearly every machine learning application can benefit from hyperparameter optimization.
The issue is that researchers often do not have a practical tool at hand. Therefore, they usually resort to manually tuning parameters. Sherpa aims to be this tool. Its goal is to require minimal learning from the user to get started. It also aims to support the user as their needs for parallel evaluation or exotic optimization algorithms grow.
As shown by references in Section~\ref{sec:motivation-and-significance}, Sherpa is already being used by researchers to achieve improvements in a variety of domains. In addition to that, the software has been downloaded more than 6000 times from the PyPi Python package manager\footnote{\url{https://pepy.tech/project/parameter-sherpa}}. It also has over 160 stars on the software hosting website GitHub. A GitHub star means that another user has added the software to a personal list for later reference.



\section{Conclusions}
\label{sec:conclusions}


Sherpa is a flexible open-source software for robust hyperparameter optimization of machine learning models. It provides the user with several interchangeable hyperparameter optimization algorithms, each of which may be useful at different stages of model development. Its interactive dashboard allows the user to monitor and analyze the results of multiple hyperparameter optimization runs in real-time. It also allows the user to see patterns in the performance of hyperparameters to judge the robustness of individual settings. Sherpa can be used on a laptop or in a distributed fashion on a cluster. In summary, rather than a black-box that spits out one hyperparameter setting, Sherpa provides the tools that a researcher needs when doing hyperparameter exploration and optimization for the development of machine learning models.

\section{Conflict of Interest}
We wish to confirm that there are no known conflicts of interest associated with this publication and there has been no significant financial support for this work that could have influenced its outcome.

\section*{Acknowledgements}
\label{sec:acknowledgements}

We would like to thank Amin Tavakoli, Christine Lee, Gregor Urban, and Siwei Chen for helping test the software and providing useful feedback, and Yuzo Kanomata for computing support. This material is based upon work supported by the National Science Foundation under grant number 1633631. We also wish to acknowledge a hardware grant from NVIDIA. 





\bibliographystyle{model1-num-names}
\bibliography{sherpa}

\pagebreak
\appendix

\section{Deep learning for Cloud Resolving Models\label{sec:appendix-case-study}}

Initially a random search was conducted on the hyperparameters listed in Table~\ref{hyperparameter-table}.

\begin{table}[H]
\footnotesize
\centering
\caption{DNN Hyperparameter Search Space.}
\label{hyperparameter-table}
\begin{tabular}{@{}ll@{}ll@{}}
\toprule
Name                        & Options    & Parameter Type         \\ \midrule
Batch Normalization\cite{ioffe2015batch}        & {[}yes, no{]}        &   Choice         \\
Dropout\cite{srivastava2014dropout, baldi2013understanding}                     & {[}0, 0.25{]}        &   Continuous  \\
Leaky ReLU coefficient\cite{agostinelli2014learning}      & {[}0 - 0.4{]}        &   Continuous      \\
Learning Rate               & {[}0.0001 - 0.01{]}  &   Continuous (log) \\
Nodes per Layer             & {[}200 - 300{]}      &   Discrete   \\
Number of layers            & {[}8 - 10{]}         &   Discrete       \\
\bottomrule
\end{tabular}
\end{table}

\begin{figure}[H]
    \centering
    \includegraphics[width=\linewidth]{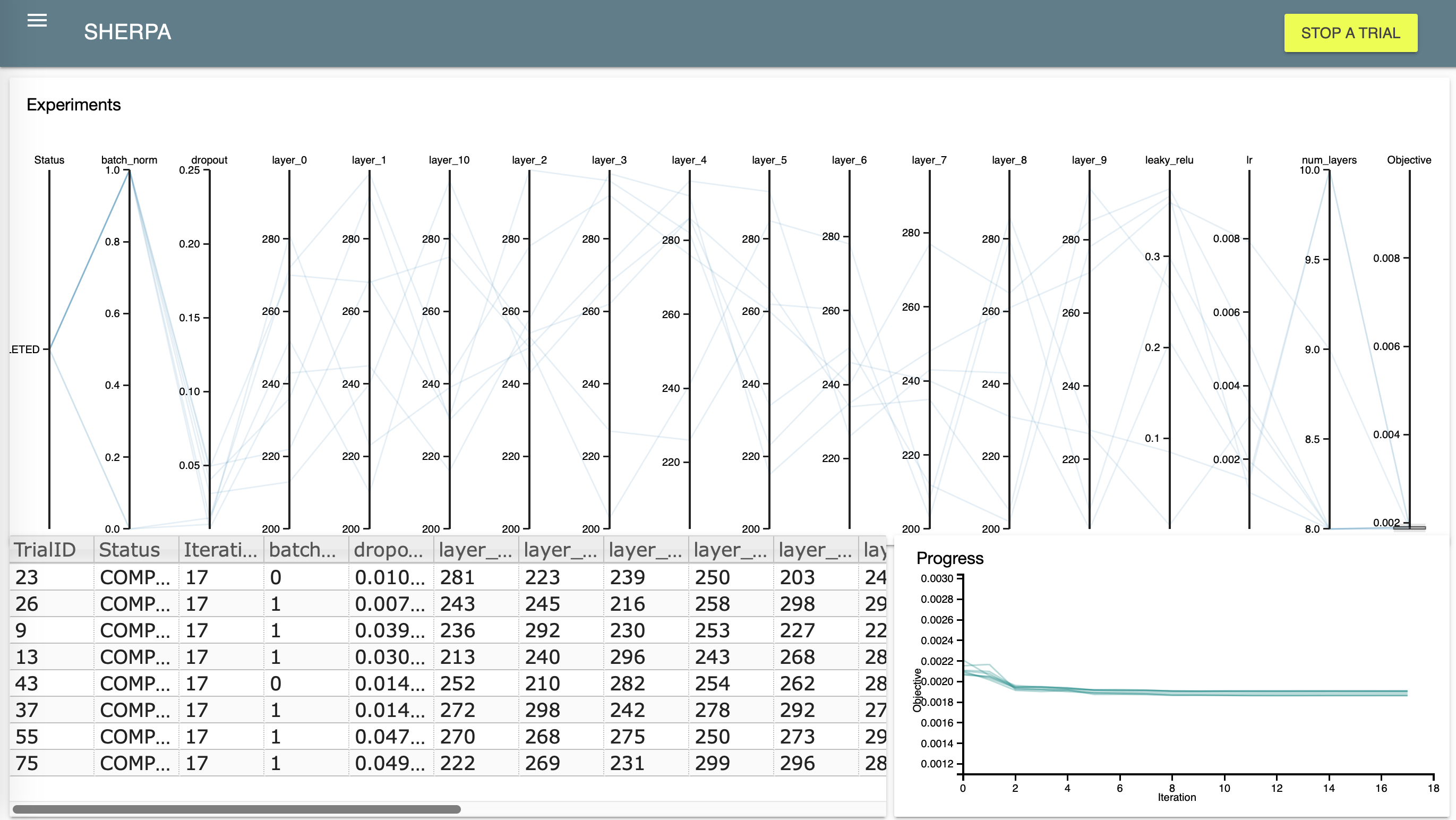}
    \caption{Screenshot of the dashboard at the end of the initial random search. The 8 best trials were selected by brushing of the \textit{Objective} axis in the parallel coordinates plot.}
    \label{fig:case_study_dashboard}
\end{figure}

A screenshot of the Sherpa dashboard at the end of the hyperparameter optimization is shown in Figure \ref{fig:case_study_dashboard} (recommended to be viewed as PDF and via zooming in). On the dashboard layer\_x refers to the number of nodes in layer $x$. From Figure~\ref{fig:case_study_dashboard} one can see that the best performing configurations have low dropout, leaky ReLU coefficients mostly around 0.3 or larger, and learning rates mostly near 0.002. The majority of good models have 8 layers and batch normalization. However, the number of units does not seem have a large impact. 

\begin{figure}[H]
\centering
\begin{subfigure}[b]{.49\linewidth}
\includegraphics[width=\linewidth]{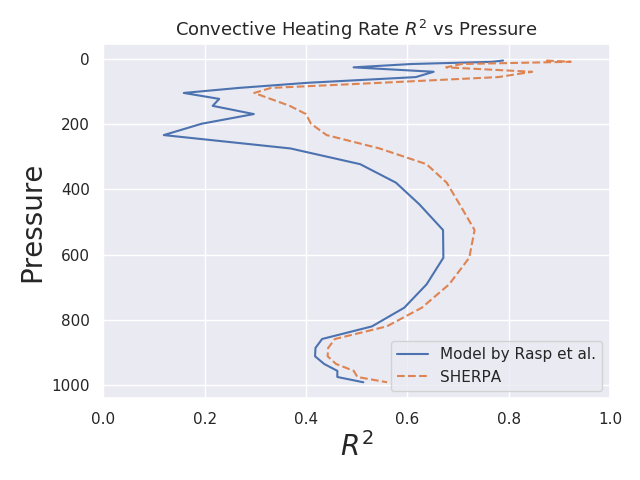}
\caption{}\label{fig:heating_rate}
\end{subfigure}
\begin{subfigure}[b]{.49\linewidth}
\includegraphics[width=\linewidth]{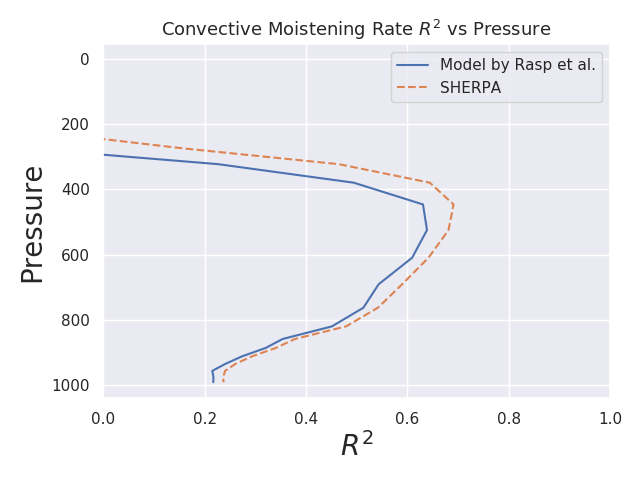}
\caption{}\label{fig:moistening_rate}
\end{subfigure}

\begin{subfigure}[b]{.49\linewidth}
\includegraphics[width=\linewidth]{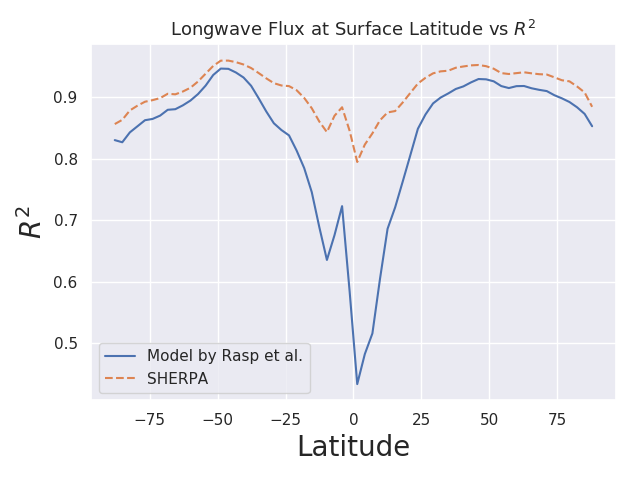}
\caption{}\label{fig:longwave_flux_at_surface}
\end{subfigure}
\begin{subfigure}[b]{.49\linewidth}
\includegraphics[width=\linewidth]{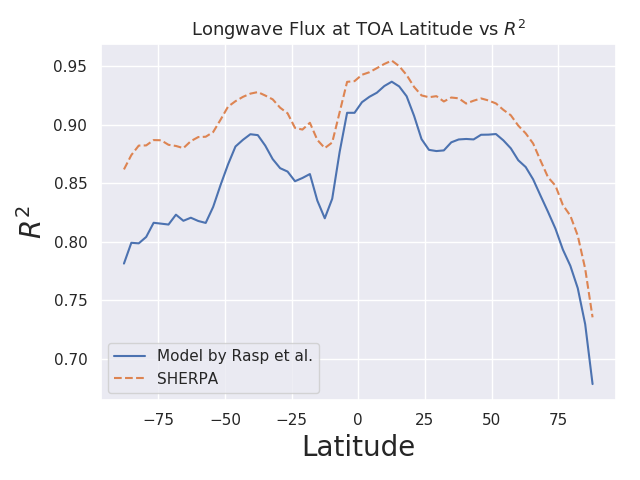}
\caption{}\label{fig:longwave_flux_at_toa}
\end{subfigure}
\begin{subfigure}[b]{.49\linewidth}
\includegraphics[width=\linewidth]{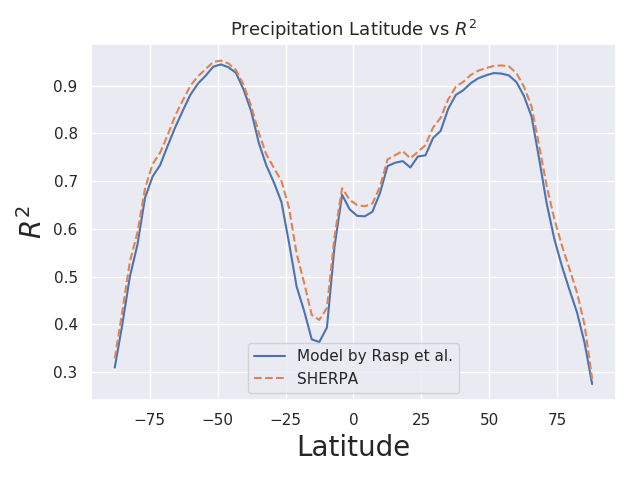}
\caption{}\label{fig:precipitation}
\end{subfigure}
\begin{subfigure}[b]{.49\linewidth}
\includegraphics[width=\linewidth]{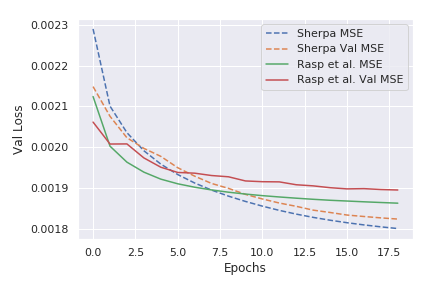}
\caption{}\label{fig:case_study_loss_plot}
\end{subfigure}
\caption{Case study results for an optimized deep neural network applied to cloud resolving models. Figures~\ref{fig:heating_rate} and~\ref{fig:moistening_rate} show the coefficient of determination $R^2$ vs. pressure for convective heating rate and convective moistening rate, respectively. Figures~\ref{fig:longwave_flux_at_surface},~\ref{fig:longwave_flux_at_toa}, and~\ref{fig:precipitation} show $R^2$ values against latitude, and~\ref{fig:case_study_loss_plot} shows loss trajectories. All figures compare the optimized Sherpa model against the model developed by \citet{rasp2018deep}.}
\label{fig:case_study_results}
\end{figure}

Following the secondary search for an optimal learning rate schedule (Section~\ref{sec:secondary-search}) the hyperparameters in Table~\ref{tab:bestconfig}) were found to be overall optimal. The optimized learning rate and schedule found by Sherpa is of considerable importance. Referencing the loss curves in Figure \ref{fig:case_study_loss_plot} one can see the learning rate schedule used in \cite{rasp2018deep} forces the learning rate to decay rapidly causing an early plateau of the loss. The learning rate schedule discovered by Sherpa on the other hand allows the DNN to keep learning, further reducing the loss.

\begin{table}[H]
\footnotesize
    \centering
    \caption{Best hyperparameter configuration found by Sherpa.}
    \begin{tabular}{ll}
    \toprule
    Batch Normalization         &   No                          \\
    Dropout                     &   0.0                           \\
    Leaky ReLU coefficient      &   0.3957                      \\
    Learning Rate               &   0.001301                    \\
    Learning Rate Decay         &   0.843784                    \\
    Nodes per Layer             &  {[}299, 269, 248, 293, 251, 281, 258, 277, 209, 270{]}\\
    Number of layers            &   10                          \\
    \bottomrule
    \end{tabular}
\label{tab:bestconfig}
\end{table}

Figure \ref{fig:case_study_results} displays results of the optimized model as they pertain to climate modeling metrics. These plots denote $R^2$ values at corresponding pressures and latitudes. Larger values of the $R^2$ indicate that the DNN is able to explain more variance in the corresponding variable. Of particular importance, are areas where Sherpa is able to perform well in regions where the previously published model fails (e.g. latitudes between -25 and 25 in Figure \ref{fig:longwave_flux_at_surface}). At all pressures and latitudes the Sherpa model outperforms the previously published model and thereby achieves a new state of the art for this dataset.

\end{document}